\documentclass[]{spie}  %>>> use for US letter paper
\usepackage[]{graphicx}

\usepackage{tikz}
\usetikzlibrary{positioning}
\usetikzlibrary{shapes,arrows,chains}
\usetikzlibrary{shapes.geometric}
\usepackage{amssymb, amsmath}
\usepackage{array, float}

\usepackage{ragged2e}
\newcolumntype{P}[1]{>{\RaggedRight\hspace{0pt}}p{#1}}

\usepackage[colorinlistoftodos,prependcaption,textsize=small,textwidth=3.5cm]{todonotes}
\newcommand*\diff{\mathop{}\!\mathrm{d}}

\newcommand{\etal}{\textit{et al}.}

\newcommand{\FT}{$\mathcal{F}_\mathcal{T}$}
\newcommand{\FS}{$\mathcal{F}_\mathcal{S}$}
\newcommand{\DT}{$\mathcal{D}_\mathcal{T}$}
\newcommand{\DS}{$\mathcal{D}_\mathcal{S}$}

\newcommand{\source}{{\mathcal S}}
\newcommand{\target}{{\mathcal T}}

\usepackage[acronym]{glossaries}

\newacronym{ADDA}{ADDA}{adversarial discriminative domain adaptation}
\newacronym{AiTR}{AiTR}{aided target recognition}
\newacronym{ATR}{ATR}{automatic target recognition}

\newacronym{BD}{BD}{Bregman Divergence}
\newacronym{BKR}{BKR}{Biomedical Knowledge Repository}
\newacronym{BRCA}{BRCA}{Breast Invasive Carcinoma}

\newacronym{CNN}{CNN}{convolutional neural network}
\newacronym{CoGAN}{CoGAN}{coupled generative adversarial nets}

\newacronym{CyCADA}{CyCADA}{cycle-consistent adversarial domain adaptation}

\newacronym{DA}{DA}{domain adaptation}
\newacronym{DAd}{DAd}{domain adversarial}

\newacronym{DAN}{DAN}{deep adaptation network}
\newacronym{DiSDAT}{DiSDAT}{direct sum domain adversarial transfer}
\newacronym{DL}{DL}{deep learning}

\newacronym{DM}{DM}{diffusion map}
\newacronym{DSDA}{DSDA}{direct sum domain adaptation}
\newacronym{DS}{DS}{direct sum}
\newacronym{EO}{EO}{electro-optical}

\newacronym{GAN}{GAN}{generative adversarial network}
\newacronym{GB}{GB}{gigabyte}
\newacronym{GPU}{GPU}{graphical processing unit}
\newacronym{GCN}{GCN}{graph convolutional network}

\newacronym{HHTL}{HHTL}{Hybrid Heterogeneous Transfer Learning}
\newacronym{HTL}{HTL}{heterogeneous transfer learning}

\newacronym{JAN}{JAN}{joint adaptation network}
\newacronym{JMMD}{JMMD}{joint maximum mean discrepancy}

\newacronym{KDE}{KDE}{kernel density estimation}

\newacronym{kNN}{k-NN}{k-nearest neighbor}
\newacronym{KL}{KL}{Kullback–Leibler}

\newacronym{LUAD}{LUAD}{Lung Adenocarcinoma}
\newacronym{MacL}{ML}{machine learning}
\newacronym{MMD}{MMD}{maximum mean discrepancy}
\newacronym{mSDA}{mSDA}{marginalized stacked denoised autoencoder}
\newacronym{MSE}{MSE}{mean square error}

\newacronym{NLM}{NLM}{National Library of Medicine}
\newacronym{NN}{NN}{neural network}

\newacronym{RAM}{RAM}{random access memory}
\newacronym{ReLU}{ReLU}{rectified linear units}
\newacronym{RKHS}{RKHS}{reproducing kernel Hilbert space}

\newacronym{SAR}{SAR}{synthetic aperture radar}
\newacronym{SGD}{SGD}{stochastic gradient descent}
\newacronym{SVM}{SVM}{support vector machine}
\newacronym{TrDM}{TrDM}{transfer diffusion map}

\newacronym{TL}{TL}{transfer learning}
\newacronym{TLDA}{TLDA}{transfer learning with deep autoencoders}
\newacronym{TSL}{TSL}{transfer subspace learning}

\makeglossaries

\title{Flexible Deep Transfer Learning by Separate Feature Embeddings and Manifold Alignment}

\author{Samuel Rivera\supit{a}, Joel Klipfel\supit{b}, and Deborah Weeks\supit{c}
\skiplinehalf
\supit{a}Matrix Research, Dayton, USA; \\
\supit{b}University of Kentucky, Lexington, USA; \\
\supit{c}George Washington University, Washington DC, USA %; \\
}

\authorinfo{Further author information: (Send correspondence to S.R.)\\
S.R.: E-mail: samuel.rivera@matrixresearch.com \\ %, Telephone: 1 937 427 8433\\
}
\begin{document}
  \maketitle

%%%%%%%%%%%%%%%%%%%%%%%%%%%%%%%%%%%%%%%%%%%%%%%%%%%%%%%%%%%%%
\begin{abstract}

Object recognition is a key enabler across industry and defense. As technology changes,
algorithms must keep pace with new requirements and data. New modalities and
higher resolution sensors should allow for increased algorithm robustness. Unfortunately,
algorithms trained on existing labeled datasets do not directly generalize to new data
because the data distributions do not match. Transfer learning (TL) or domain adaptation
(DA) methods have established the groundwork for transferring knowledge from existing
labeled source data to new unlabeled target datasets.
However, current DA approaches assume similar source and target feature spaces
and suffer in the case of massive domain shifts or changes in the feature space. Existing
methods assume the data are either the same modality, or can be aligned to a
common feature space. Therefore, most methods are not designed to support a
fundamental domain change such as visual to auditory data.

We propose a novel deep learning framework that overcomes this limitation by learning
separate feature extractions for each domain while minimizing the distance between the domains in a latent lower-dimensional space. The alignment is achieved by considering
the data manifold along with an  adversarial training procedure. We demonstrate the
effectiveness of the approach versus traditional methods with several ablation experiments
on synthetic, measured, and satellite image datasets. We also provide practical guidelines for training the network while overcoming vanishing gradients which inhibit learning in some adversarial training settings.

\end{abstract}

%>>>> Include a list of keywords after the abstract

\keywords{transfer learning, domain adaptation, adversarial learning, deep learning, machine learning, automatic target recognition, classification}

%%%%%%%%%%%%%%%%%%%%%%%%%%%%%%%%%%%%%%%%%%%%%%%%%%%%%%%%%%%%%
\section{Introduction}
\label{sec:intro}  % \label{} allows reference to this section
Object recognition, the process of using machines to classify objects from sensor data, is a key enabler across industry and defense that supports automation and situational
awareness. \Gls{DL} algorithms have achieved state-of-the-art performance on current
vision tasks by learning good feature representations for large labeled datasets
\cite{Simonyan15}. Labeling datasets relevant to the sensor domain, however, is often
prohibitively expensive due to the abundance of new data available for emerging
sensor modalities, and the cost of collecting and labeling appropriate data for all
relevant operating conditions (environment, target, and sensor state). Using similar, labeled datasets
or generating synthetic datasets mitigates some of these issues, but typically fails to generalize well. Namely, classification performance on the \emph{target} dataset
of interest suffers when the \emph{source} dataset used for training has a much different
probability distribution \cite{Pan2010}.

\begin{figure}
   \begin{center}
   \begin{tabular}{c}
   \includegraphics[width=.3\textwidth]{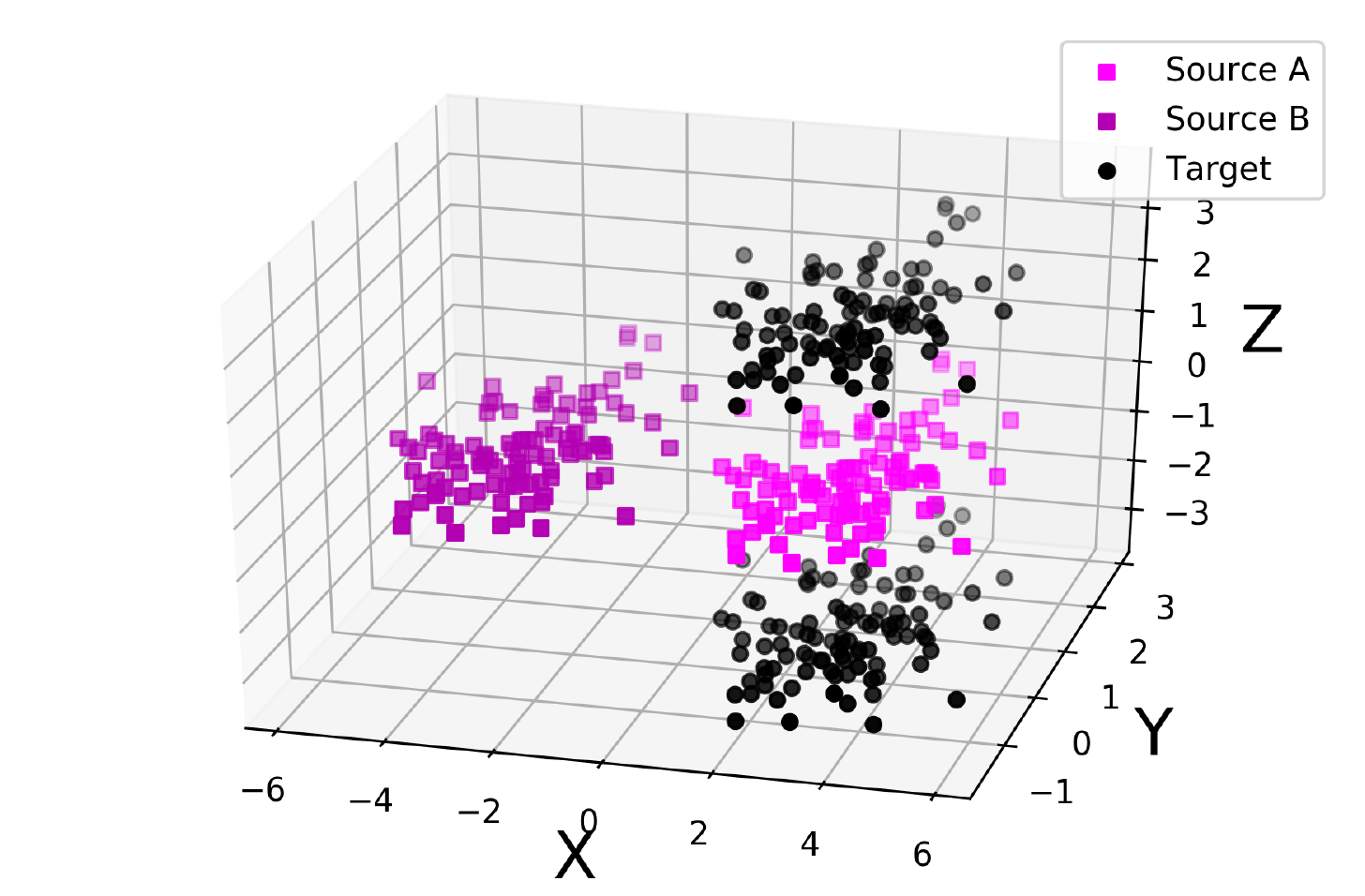}
   \end{tabular}
   \end{center}
   \caption[example]{ \label{fig:2Dto3D} Illustration of a cross-feature-space transfer challenge. }
\end{figure}

\Gls{TL} or \gls{DA} methods have established the groundwork for transferring knowledge
from existing labeled source data to these new unlabeled target datasets to overcome the
burden of requiring labeled datasets. Unfortunately, most \gls{DA} approaches assume
similar source and target feature spaces and suffer in the case of massive domain shifts
or are not meant to handle a fundamental change in the feature space. Therefore, most
existing methods assume the data are either the same modality, or can be aligned to a
common feature space. We illustrate the challenge in Fig.~\ref{fig:2Dto3D}. In this example,
the source classes A and B form distinct clusters on the $xy$-plane that are separated at
the origin by the $yz$-plane. The target data also form two distinct clusters, but are
separated by the $xy$-plane and completely overlap with source class B of projects onto
the $xy$-plane. This example shows a case where the  discriminative feature information
from the source data (x values are informative) has no bearing on category identity for
the target samples.

This challenge of learning to transfer across fundamentally different features spaces, or
where discriminative information from one domain is not helpful for the other, is referred
to as \gls{HTL}, and it has received less attention than its \gls{TL} counterpart. See Day
and Khoshgoftaar \cite{Day2017} for a recent survey of this approach which has received
relatively less attention within the \gls{DL} literature. We propose a novel deep learning
algorithm called \gls{DiSDAT} that can transfer across very different domains by learning
separate feature extractions for each domain while minimizing the distance between the two
domains in a latent lower-dimensional space. A key idea behind the \gls{DiSDAT} network is
that source and target feature spaces may be fundamentally different so source and target
inputs should be modeled as existing in separate feature spaces as opposed to assuming a
common input feature space.

\begin{figure}[h]
   \begin{center}
   \begin{tabular}{c}
   \includegraphics[width=.7\textwidth]{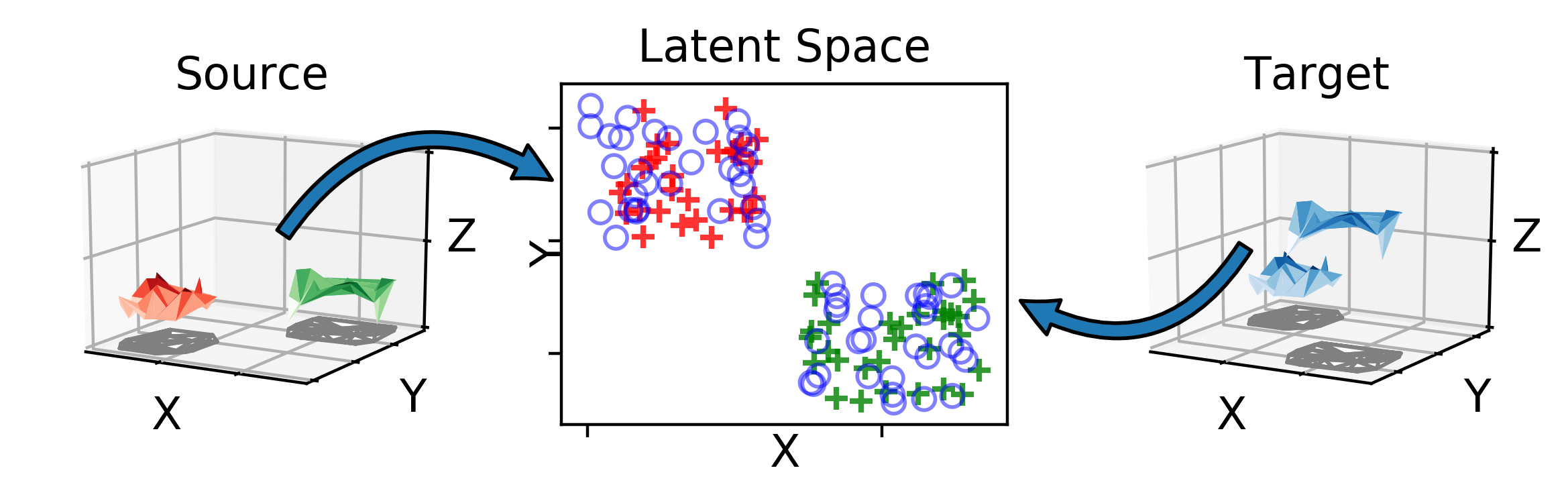}
   \end{tabular}
   \end{center}
   \caption[Direct Sum Illustration]{ \label{fig:dirSumIllustration} Illustration of the projection from two different feature spaces to a common latent space.}
\end{figure}

A key assumption is that the source and target data, although residing in different feature
spaces, exist on lower dimensional manifolds where the probability distributions are
isomorphic, or can be aligned by some transformation of the original feature spaces.
Therefore, we align the source and target by learning a feature extraction that preserves
the underlying manifold while minimizing the divergence between the source and target
distributions in the latent space as illustrated in Fig.~\ref{fig:dirSumIllustration}.
We accomplish this by define a suitable objective function that penalizes divergence
between the source and target distributions in the latent space. Once aligned in a latent
space, we readily classify target data using a classifier trained on labeled source data.

The paper is organized as follows: We describe related work in Sec.~\ref{sec:related} before providing \gls{DiSDAT} details in Sec.~\ref{sec:methods}. We demonstrate the effectiveness of the approach versus traditional methods with several ablation experiments on synthetic, measured, and satellite image datasets in Sec.~\ref{sec:experiments}. We also provide practical guidelines for training the network while overcoming vanishing gradients which inhibit learning in some adversarial training settings. Results will show that incorporating a generative manifold-preserving framework within the network architecture alleviates degenerate learning that is often resolved by initializing the target feature extractor with the source extractor\cite{tzeng2017adversarial}. Finally, we close with a discussion and conclusions.

\section{Related Work}
\label{sec:related}
%
% - denote the specific sub-area where we exist (reference in terms of the terminology used by the seminal reference paper).
%     - Unsupervised transfer
%     - Access to unlabeled target data
%
% More recently, Deep learning has taken note (cite Deep transfer)
%     - Summarize the key high-level ideas (sub-approaches)
%     - note the sub-area where we fit
%
% We focus on very relevant work
% - manifold learning/alignment/regularizing
% - separate feature embeddings
% - divergence minimization

\Gls{TL} remains an active area of research because of the important
applications described above. See Pan \& Yang \cite{Pan2010} for a thorough review of
\gls{TL} across classification, clustering, and regression methods. Our work falls
under the subcategory Pan \& Yang refer to as \emph{transductive transfer learning},
where the source and target tasks are the same but over different domains. Furthermore,
we consider the case where labels exist for the source data and the algorithm has access
to only unlabeled target data. This type of transfer, where we want to learn a
classifier from source data but apply it to a different target dataset is also called
\gls{DA}, which has been a very active area of research recently within \gls{DL} for
visual applications \cite{Csurka2017,WangSurvey2018}. In this section we do not attempt
to cover all avenues of work, but instead refer to relevant review papers for context
and summarize the most relevant work here.

One highly relevant method by Tzeng and colleagues, called \gls{ADDA}
\cite{tzeng2017adversarial}, establishes a framework for \gls{DA} where separate source
and target feature extraction mappings are learned that minimize the \emph{distance}
between source and target while being discriminative for the source data. A classifier
learned with the source data then extends to target data after the mapping step. This
general framework allows for choice of i) whether source and target feature extractions
should share weights, ii) similarity (or divergence) measure between source and target,
and iii) whether the algorithm employs a generative or purely discriminative
architecture. Importantly, many of the recent \gls{DA} methods can be described as a
variant of this framework. Our approach also fits within this general framework of
aligning the source and target in a discriminative latent feature space, but with
important algorithmic distinctions for each major component. Therefore, we describe the
key related literature in the context of the \Gls{ADDA} framework.

{\bf Network Weight Sharing:}
A major drawback of \gls{ADDA} and related methods is that they use the source mapping
parameters to initialize the target mapping parameters, then fine-tune the target
mapping parameters\cite{Yosinski2014} to prevent the network from learning a degenerate
target mapping. This means that source and target must be pre-aligned to the same feature
space. In our early experiments we also found that learning a separate target mapping
when using a domain adversarial loss\cite{Ganin15MLR} for source and target alignment
led to poor target mapping solutions when starting with random initial weights. We overcame this by using an autoencoder
to preserve the underlying structure of the original target data while learning the
mappings. This allows us to learn a target mapping from random initializations and remove
the requirement of having mirrored network architectures for source and target feature
mapping. Consequently, this allows for much larger domain shifts and completely different
modality shifts, such as in \gls{HTL}, that would not be possible by initializing with
the source mapping because the target mapping does not depend on the source feature mapping.

A notable example of \gls{HTL} within the \gls{DL} family is the \gls{HHTL} algorithm of
Zhou \etal\cite{Zhou2014}. \Gls{HHTL} first learns separate higher-level representations
of source and target data using \gls{mSDA}\cite{Chen2012}, a particular variant of
auto-encoders that learn increasingly higher-level data representations. Then, target
samples in the latent space are mapped to the source latent space where a source
classifier applies across both source and target. Unfortunately, learning the mapping
between source and target latent representations requires corresponding samples between
source and target, which limits the practical application of \gls{HHTL}. Our method does
not require corresponding samples between source and target, so it is more applicable in
practice.

{\bf Divergence Measure:}
Another important consideration within the general \gls{DA} framework is the choice of
divergence measure between source and target as well as the training procedure.
\Gls{ADDA} uses a \emph{GAN-loss}, a domain classifier loss with reversed labels, to
adapt the target to the source in the latent space. The GAN-loss builds on the seminal
work of Ganin \etal\cite{Ganin15MLR} that uses the domain classification loss along with
\emph{gradient reversal} to maximize the domain classification error, essentially learning
feature mappings such that the source and target are indistinguishable in the latent space.
Using the GAN-loss instead of the gradient reversal achieves the same goal but avoids vanishing gradients that occur in practice when training \gls{DA} networks using
the gradient reversal trick\cite{tzeng2017adversarial}.

Unfortunately, the domain classification error only acts as a \emph{proxy} for the
similarity between the source and target distributions in the latent space and does not
explicitly measure the distance between the source and target distributions. An
alternative discrepancy measure, the \gls{MMD}\cite{Borgwardt2006,Gretton2012}, more
directly measures the difference between source and target distributions through the distance in a \gls{RKHS}. Long
\etal\cite{Long2015,Long2018} applied \gls{MMD} as a key component of the \gls{DAN}
algorithm for minimizing the discrepancy between source and target in higher level
network activation layers. The idea has been extended to the \gls{JMMD} in
\gls{JAN}\cite{Long2017} to better minimize the joint discrepancy across higher level
layers and lower level feature extraction layers.

Taking inspiration from the subspace transfer learning work of Si Si \etal\cite{SiSi2010}
and Mendoza \etal\cite{Mendoza2017}, we use a quadratic form of
\gls{BD}\cite{Bregman1967} as the measure of dissimilarity between source and target
distributions in the latent space. This allows us to explicitly model differences
between source and target distributions in the latent space and learn feature mappings
such that source and target distributions match. Unlike \gls{DAN}, we do not try to
match network outputs for source and target across the network, but instead focus on
aligning distributions after feature extraction. Once source and target
data are aligned in a common latent feature space, then the higher classification layers
do not need to be further altered for the target.

Other researchers have used the \gls{KL}-divergence\cite{Kullback51}, another special
case of \gls{BD}, to penalize the divergence between two vectors that represent
probability mass functions. This type of penalty has been used as an alternative to the
standard cross-entropy loss for multi-class classification. It has also been applied by
Zhuang \etal~in the \gls{TLDA} algorithm\cite{zhuang2015supervised} to penalize the
distance between source and target in the embedded space. A major shortcoming of their
method is that the distributions are not actually aligned. Instead, the difference
between the \emph{sample means} of source and target in the embedded space are penalized without penalizing the difference between the overall distributions.
\Gls{TLDA} also deserves mention because it applies an autoencoder similarly to our
algorithm, but unlike our algorithm assumes tied input source and target feature mapping as with most algorithms.

{\bf Architecture:}
The final point of discussion is the specific network architecture. Tzeng
\etal\cite{tzeng2017adversarial} frame the discussion in terms of whether networks are
purely discriminative, or whether they include a generative modeling component.
Generative models typically use a variant of \glspl{GAN}\cite{Goodfellow2014} or
autoencoders. Tzeng and colleagues argue that it is not always necessary to learn a
generative data model when learning a discriminative model for classification. We
sympathize with this view, but acknowledge that cases exist where generative models can
support discriminate model learning. Sankaranarayanan \etal
\cite{Sankaranarayanan2018}, for example, use a \gls{GAN} along with the learned
embedding for \gls{DA}. They argue that learning the generative model produces more
meaningful gradients for backpropagation.

\Glspl{GAN} have also been applied in the \gls{CoGAN} algorithm\cite{Liu2016}
to learn shared high-level representations that are common across domains. This is
similar to the idea of using \gls{mSDA} to learn higher level representations of data
\cite{Chen2012} that work across domains. Alternatively, \Glspl{GAN} can be used along
with \emph{cycle consistency} as in \gls{CyCADA}\cite{Hoffman18a} to adapt data across
domains such that both feature-level and higher-level semantic information is preserved.

Besides learning high-level representations, a generative model can learn the
underlying data structure, or manifold. The approach by Gong \etal~projects the source
and target in a Grassmann manifold, then aligns the data using the
\emph{geodesic flow kernel} approach\cite{Gong2012}. Our approach is similar to this in
that it learns and aligns the underlying source and target data manifolds, but the
manifold is learned through an autoencoder and the alignment is achieved by divergence
minimization. Others have used \gls{GCN}\cite{kipf2017} to model the underlying data
structure for semi-supervised \gls{DL}.

\section{METHODS}
\label{sec:methods}  % \label{} allows reference to this section

\subsection{Direct Sum Domain Adversarial Transfer Network (DiSDAT)}\label{sec:dsda}  % \label{} allows reference to this section
\begin{figure}
   \begin{center}
   \begin{tabular}{c}
    % -------------------------------------------------
% Set up a new layer for the debugging marks, and make sure it is on
% top
\pgfdeclarelayer{marx}
\pgfsetlayers{main,marx}
% A macro for marking coordinates (specific to the coordinate naming
% scheme used here). Swap the following 2 definitions to deactivate
% marks.
\providecommand{\cmark}[2][]{%
  \begin{pgfonlayer}{marx}
    \node [nmark] at (c#2#1) {#2};
  \end{pgfonlayer}{marx}
  }
\providecommand{\cmark}[2][]{\relax}
% -------------------------------------------------
% Start the picture
\begin{tikzpicture}[%
    >=triangle 60,              % Nice arrows; your taste may be different
    start chain=going below,    % General flow is top-to-bottom
    node distance=5mm and 30mm, % Global setup of box spacing
    every join/.style={norm},   % Default linetype for connecting boxes
    ]
% -------------------------------------------------
% A few box styles
% <on chain> *and* <on grid> reduce the need for manual relative
% positioning of nodes
\tikzset{
  base/.style={draw, on chain, on grid, align=center, minimum height=4ex,minimum width=15mm},
  proc/.style={base, rectangle}, % text width=25mm}, %8em},
  test/.style={base, diamond, aspect=2, text width=5em},
  term/.style={proc, rounded corners},
  % coord node style is used for placing corners of connecting lines
  coord/.style={coordinate, on chain, on grid, node distance=6mm and 15mm },
  % nmark node style is used for coordinate debugging marks
  nmark/.style={draw, cyan, circle, font={\sffamily\bfseries}},
  %-------------------------------
  % rivera defined styles
  input/.style={base, rectangle, rounded corners  }, %, fill=blue!30},
  output/.style={base, rectangle}, % fill=green!30},
  encode/.style={base, trapezium, trapezium angle=-50}, %, fill=orange!20},
  decode/.style={base, trapezium, trapezium angle=50}, %, fill=orange!20},
  % -------------------------------------------------
  % Connector line styles for different parts of the diagram
  norm/.style={->, draw}, %, blue!30}, % lcnorm},
  free/.style={->, draw, red!30}, %lcfree},
  cong/.style={->, draw, green!30}, %lccong},
  it/.style={font={\small\itshape}}
}

% -------------------------------------------------
% % Rivera putting down nodes
\node [input](start) {Source Input};
\node [encode,join](fs) {\FS};
\node [coord,join] (mergeLeft){};
\node [decode,join](ds) {\DS};

\node [proc,right=of mergeLeft,xshift=-2.6cm] (merge) {Latent Space};
\node [output,join,below=2.5cm of merge,xshift=-1cm](classifier) {Classifier};
\node [input,join](classEnd) {Class\\Output};

\node [output,below=1.58cm of merge,xshift=1.5cm](gradReverse) {Gradient Reversal};
\node [output,join,yshift=.2cm](domainClassifier) {Domain Classifier};
\node [input,join](domainEnd) {Domain\\Output};

\node [input,right=of start] {Target Input};
\node [encode,join](ft) {\FT};
\node [coord,join] (mergeRight){};
\node [decode,join](dt) {\DT};

% arrows
\path (mergeLeft.east) to node [near start, xshift=0em] {} (merge);
  \draw [->] (mergeLeft.east) -- (merge);
\path (mergeRight.west) to node [near start, xshift=0em] {} (merge);
\draw [->] (mergeRight.west) -- (merge);

\path (merge) to node [near start, xshift=0em] {} (gradReverse);
\draw [->] (merge) -- (gradReverse.north west);

\end{tikzpicture}
   \end{tabular}
   \end{center}
   \label{fig:DiSDAT}
   \caption[example]{ \label{fig:network} Illustration of the \gls{DiSDAT} Network.}
\end{figure}
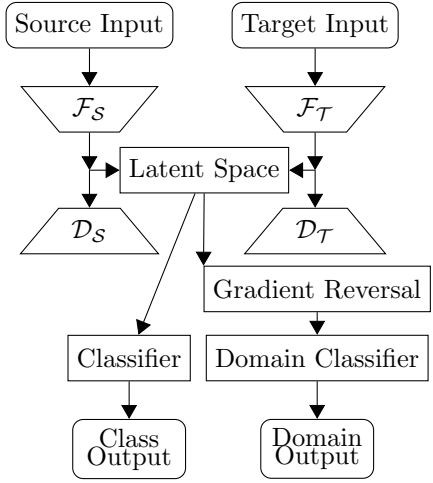

We model the input feature space mathematically as the algebraic sum of the source and
target feature spaces. More formally the source space $\mathcal{S}$ and target space
$\mathcal{T}$ give rise to a combined input space, $\mathcal{S}\bigoplus\mathcal{T}$.
Separate feature extraction modules (\FS,~\FT) embed data from either $\mathcal{S}$ or
$\mathcal{T}$ into a common latent feature space, $\mathcal{H}$. Separate feature
embeddings allow the source and target to differ
arbitrarily--even in dimensionality--as long as the feature embeddings preserve isomorphisms
between the class conditional probability distributions of the source and target data.
As indicated in Figure \ref{fig:DiSDAT}, DiSDAT learns the latent feature space embeddings
\FS, ~\FT~ using separate auto-encoders.
The auto-encoder used to train \FS~(or \FT)
first maps from $\mathcal{S}$ (or $\mathcal{T}$) into the common latent feature space,
and then back out to $\mathcal{S}$ (or $\mathcal{T}$) with the decoders $\mathcal{D}_\mathcal{S}$ ( or $\mathcal{D}_\mathcal{T}$).
We learn the embedding
\FS (or \FT) from $\mathcal{S}$
(or $\mathcal{T}$) into the common latent feature space  and the corresponding decoder by applying standard backprop procedure with the mean square error (MSE) of the reconstruction loss  between the input and output of
an autoencoder. Using an autoencoder to train the feature extraction modules
\FS~and \FT~allows DiSDAT to account for
the underlying geometric structure of $\source$ and $\target$, when considered
as Grassmannian manifolds. Once features are embedded in a common latent space, \gls{DiSDAT} assigns to a category label using a classifier $\mathcal C$ that is trained using backprop updates along with the standard cross-entropy loss.

% Once mapped to a common hidden representation, the classifier function $\mathcal{C}$ maps
% samples to a class label.

The key feature of DiSDAT is that it learns an embedding into the common latent feature space $\mathcal{H}$ that aligns the source and target data probability distributions. DiSDAT does so by employing the quadratic-type Bregman divergence
discussed in Si \etal\cite{SiSi2010} and Mendoza \etal\cite{Mendoza2017} to measure
differences between the respective probability distributions $P_\source$ and $P_\target$ for the
embedded source and target data, and then updates the source and target data via gradient descent.
The quadratic-type Bregman divergence from Si \etal\cite{SiSi2010} and Mendoza \etal\cite{Mendoza2017}
is given by
\begin{align} \label{eq:breg_div}
	D(P_\source\| P_\target) = \int(P_\source({\bf y}) - P_\target(\bf y))^2 \, \diff{\bf y},
\end{align}
where the above integral is taken over the common latent feature space. By using
Gaussian \gls{KDE}, Si shows that \eqref{eq:breg_div} can
be approximated on the embedded source and target data by
\begin{align} \label{eq:breg_approx}
	D(P_\source \| P_\target)
		&\approx \frac{1}{(n_\source)^2} \sum_{j = 1}^{n_\source} \sum_{k=1}^{n_\source}
			G_{\Sigma_{\source,\source}}({\bf y}^\source_k - {\bf y}^\source_j) \nonumber\\
		&\quad + \frac{1}{(n_\target)^2} \sum_{j = 1}^{n_\target} \sum_{k = 1}^{n_\target}
			G_{\Sigma_{\target,\target}}({\bf y}^\target_k - {\bf y}^\target_j) \nonumber\\
		&\quad - \frac{2}{n_\source \, n_\target} \sum_{j = 1}^{n_\source} \sum_{k = 1}^{n_\target}
			G_{\Sigma_{\source,\target}}({\bf y}^\target_k - {\bf y}^\source_j).
\end{align}
In Equation \eqref{eq:breg_approx}, $\{{\bf y}^\source_k\}_{k=1}^{n_\source}$ and
$\{{\bf y}^\target_k\}_{k=1}^{n_\target}$ are respectively used to denote the
embedded source and target data, where $n_\source$ and $n_\target$ are the respective
cardinalities of the source and target datasets. We further use
$G_{\Sigma_*}$ to denote the Gaussian kernel
\[
	G_{\Sigma_*}({\bf y})
		= \exp\left(-\frac{1}{2}{\bf y}^T \Sigma_*^{-1} {\bf y} \right)
\]
with covariance matrix $\Sigma_*$, where
$\Sigma_\source$ and $\Sigma_\target$ respectively represents the covariance matrices
for the embedded source and target data, and the notation ${\bf y}^T$ denotes
the transpose vector ${\bf y}$. In order to simplify notation,
we follow Si's lead and define $\Sigma_{*,*} := \Sigma_* + \Sigma_*$.
For example, under this notational convention, $\Sigma_{\source, \target}$ represents
the sum $\Sigma_{\source, \target} = \Sigma_\source + \Sigma_\target$. The loss function $L_{\text{Breg}}$
used in DiSDAT to penalize for differences data probability distributions
is precisely \eqref{eq:breg_approx}. To perform backpropagation, one can
readily show that the derivatives of $D(P_\source \| P_\target)$ with respect
to the embedded source and target data are given by
\begin{align}\label{breg-partial_s}
	\frac{\partial}{\partial {\bf y}^\source_i} D(P_\source\|P_\target)
		&=\frac{2}{(n_\source)^2} \sum_{k=1}^{n_\source} G_{\Sigma_{\source, \source}}
			({\bf y}^\source_k - {\bf y}^\source_i)
				\left(\Sigma_{\source, \source}\right)^{-1} ({\bf y}^\target_k - {\bf y}^\target_i)
			\nonumber \\
		&\qquad -\frac{2}{n_\source \, n_\target} \sum_{k=1}^{n_\target}
			G_{\Sigma_{\source, \target}}({\bf y}^\target_k - {\bf y}^\source_i)
				\left(\Sigma_{\source, \target}\right)^{-1}({\bf y}^\target_k - {\bf y}^\source_i)
\end{align}
and
\begin{align}\label{breg-partial_t}
	\frac{\partial}{\partial {\bf y}^\target_i} D(P_\source\|P_\target)
		&=\frac{2}{(n_\target)^2} \sum_{k=1}^{n_\target} G_{\Sigma_{\target, \target}} ({\bf y}^\target_k - {\bf y}^\target_i)
				\left(\Sigma_{\source, \source}\right)^{-1} ({\bf y}^\target_k - {\bf y}^\target_i)
			\nonumber \\
		&\qquad -\frac{2}{n_\source \, n_\target} \sum_{k=1}^{n_\source} G_{\Sigma_{\source, \target}}({\bf y}^\source_k - {\bf y}^\target_i)
				\left(\Sigma_{\source, \target}\right)^{-1}({\bf y}^\target_k - {\bf y}^\source_i).
\end{align}

An alternative approach to divergence minimization for \gls{DA} is the \gls{DAd} technique of Ganin \etal\cite{Ganin15}, where the network is optimized to \emph{increase} the domain classification loss by using gradient descent on the reversed domain classifier loss. As noted above, \gls{DAd} is distinct from \gls{BD} regularization as \gls{DAd} does not
align probability distributions, but instead uses the domain classifier loss as a proxy for distribution divergence. To compare this popular approach to \gls{BD}, we implemented \gls{DAd} within our framework using the standard cross-entropy loss and a gradient reversal layer approach\cite{Ganin15} after the embedding layer as shown in Fig.~\ref{fig:DiSDAT}. In following section we will describe how to toggle between the regularizations by altering the total cost function.

% DiSDAT incorporates Inspired by the work of Ganin, Ustinova, {\em et al.,}\cite{Ganin15}

\subsection{Network Sculpting Through Loss Customization}
To understand the effect of the different features of \gls{DiSDAT}, we performed a series of ablation studies on various transfer tasks and datasets. Each
study consists of 12 separate experiments which are based on
modifying the parameters of the \gls{DiSDAT} loss function:
\begin{align}
	\label{eq:loss_func}
	L = \lambda_{\text{AE}_\source} \, L_{\text{AE}_\source}
		+ \lambda_{\text{AE}_\target} \, L_{\text{AE}_\target}
		+ \lambda_{\text{class}} \, L_{\text{class}}
		- \alpha_{\text{DA}} \, L_{\text{DA}_\source}
		-\alpha_{\text{DA}} \, L_{\text{DA}_\target}
		+ \lambda_{\text{Breg}} \, L_{\text{Breg}}
\end{align}
Here we denote the source and target space auto-encoder loss by $L_{\text{AE}_*}$, the
 classifier loss by $L_{\text{class}}$, the \gls{DAd} loss by
$L_{\text{DA}_{*}}$, and the \gls{BD} loss by $L_{\text{Breg}}$. By setting particular loss parameter values,
we can control the specific network components involved during network optimization. The parameter values
for each experiment type are shown in Table \ref{tab:exp-abla}. Note that the final column of Table
\ref{tab:exp-abla} indicates whether or not a direct sum is used to
separately embed $\source$ and $\target$ into the common latent feature space.
\begin{table}[H]
	\centering
	\begin{tabular}{P{4cm}cccccc}
		Experiment & $\alpha_{\text{DA}}$ & $\lambda_{\text{AE}_\source}$ & $\lambda_{\text{AE}_\target}$ & $\lambda_{\text{class}}$ & $\lambda_{\text{Breg}}$ & Sep Embedding \\
		\hline
		Baseline                   & 0   & 0 & 0 & 1 & 0 & No \\
		Domain Adversarial (DA)    & 0.1 & 0 & 0 & 1 & 0 & No \\
		Bregman Divergence(BD)     & 0   & 0 & 0 & 1 & 1 & No \\
		Auto-Encoder (AE)          & 0   & 1 & 1 & 1 & 0 & No \\
		DA, AE                     & 0.1 & 1 & 1 & 1 & 0 & No \\
		BD, AE                     & 0   & 1 & 1 & 1 & 1 & No \\
		Direct Sum (DS)            & 0   & 0 & 0 & 1 & 0 & Yes \\
		DS, DA                     & 0.1 & 0 & 0 & 1 & 0 & Yes \\
		DS, BD                     & 0   & 0 & 0 & 1 & 1 & Yes \\
		DS, DA, AE                 & 0.1 & 1 & 1 & 1 & 0 & Yes \\
		BD, BS, AE                 & 0   & 1 & 1 & 1 & 1 & Yes \\
		Everything                 & 0.1 & 1 & 1 & 1 & 1 & Yes \\
	\end{tabular}
	\caption{Experiment Types for Ablation Study}
	\label{tab:exp-abla}
\end{table}

\section{EXPERIMENTS}
\label{sec:experiments}  % \label{} allows reference to this section

% \begin{figure}
% 	\centering
% 	\includegraphics{Fig/NetworkDesign}
% 	\caption{Netowrk Design}
% 	\label{fig:network}
% \end{figure}

\subsection{Network Implementation Details}
We used a small convolutional network to establish a proof of concept for \gls{DiSDAT}. Source and target feature encoders (\FS~ and \FT) began with a 2D convolution layer having 1 input channel (grayscale images), 16 output channels, a 3 $\times$ 3 kernel, a stride size of 1, and  1 pixel padding. This was fallowing by 2D batch normalization, a ReLU activation, and 2D max pooling with a 2$\times$ 2 kernel a stride of 2. Next, we applied a 2nd 2D convolution with 16 input channels, 8 output channels, a 3 $\times$ 3 kernel, a stride of 2 and 1 pixel padding. The second convolution was also followed with batch normalization, ReLU activation, and max pooling but with a stride of 1 pixel. We closed the feature extraction with a fully connected layer having 32 inputs and either 3 or 10 outputs depending on the experiment.

The decoders (\DS~ and \DT) reversed the convolution procedure, with slight adjustments.  We began with a fully connected layer with 32 output channels. Then we applied a 2D transpose convolution operation with 8 input channels, 16 output channels, a 3 $\times$ 3 kernel, a stride size of 2 and no padding. This was followed by ReLU activation and 2D batch normalization before another transpose convolution. The 2nd transpose convolution had 16 input channels, 8 output channels, a 5$\times$5 kernel, a stride size of 3, and a 1 pixel padding. Next we applied ReLU activation and batch normalization. The final 2D transpose convolution had 8 input channels, 1 output channel, a 2 $\times$ 2 kernel, a stride of 2, and 1 pixel padding. We finally applied the hyperbolic tangent.

The classifier network operated on the output of the feature encoders (\FS~ and \FT). We applied a fully connected layer with 5 output nodes followed by ReLU activation and a final fully connected layer with a number of output nodes equal to the number of classes. The domain classifier was a single fully connected layer with 2 output nodes.

For the \gls{BD} implementation, parameter choices for the \gls{KDE} have important consequences regarding accuracy and computation. For this implementation we made a simplifying assumption that latent features are independent, which leads to diagonal covariance matrices that are estimated by simply calculating the feature variances. This leads to trivial inverse covariance matrix estimation by inverting the diagonal values. For numerical reasons, we found it necessary to apply Tikhonov regularization\cite{Tikhonov} for matrix inversion by adding $0.001$ to the variance values before inverting.

\subsection{Experiment Setup}
Both cross class and cross dataset transfer tasks were used to validate the \gls{DiSDAT} architecture. For cross class transfer tasks, we took disjoint two class subsets from the same dataset as the source and target domains. Cross dataset transfer tasks involving taking two distinct datasets as source and target domains.

Each experiment for every architecture evaluated was performed
using 5 Monte Carlo iterations. Reported results are the average accuracy over the Monte Carlo iterations, and the reported errors for each experiment are the standard deviations across all Monte Carlo iterations. For the two cross-class experiments, it is possible that the transfer network reversed the target class labels. That is to say that the transfer network will assign a 0 or 1 arbitrarily to either of the two possible class options. Therefore, for two-class transfer experiments we transformed accuracy as $\max(\text{accuracy}, 1-\text{accuracy})$ for each Monte Carlo trial to account for a possible label reversal.

\subsection{Databases}
The transfer tasks we use to benchmark DiSDAT employ the Fashion-MNIST (FMNIST)\cite{FMNIST}, MNIST\cite{MNIST}, USPS\cite{USPS}, xView\cite{xView}, and SAMPLE\cite{SAMPLE} datasets.
Both MNIST and USPS consist of gray-scale images of the handwritten digits 0 through 9 that we scaled to 28 $\times$ 28 pixels. The MNIST training set has 60,000 images that were downsampled to a common 7,000 images across our experiments. The USPS training set contains 7,291 images that were downsampled to 7,000 images for our experiments.

Due to the ease with which convolutional neural networks can classify the MNIST images, FMNIST was designed by Zalando to allow researchers to better showcase improvements in neural network architectures by providing researchers with a dataset that is more difficult for networks to classify. FMNIST consists of 28 $\times$ 28 pixel grayscale images of articles of
clothing and footwear from ten distinct classes. The FMINST training set has 60,000 images that are evenly split over the 10 classes. Particular subsets were used in our experiments. The 10,000 test images were not used.

xView is a particularly large dataset with approximately 1.0 million satellite images taken at 0.3 meter resolution. The object training images in xView are annotated using bounding boxes, and are labeled according to 60 distinct classes. We generated our xView images by cropping square regions centered at the bounding box locations then scaling to 28 $\times$ 28 pixels. We used a subset of 1000 images from each class of interest across our experiments. A detailed description of the xView dataset is found in a paper by Darius Lam \etal\cite{xView}

The SAMPLE dataset is a publicly available synthetic and measured SAR imagery dataset of 10 target classes. This set is particularly small, with just 1345 total images across all classes in the synthetic set and 1345 in the measured set. We used the \emph{decibel} format images for our experiments.

\subsection{Exp 1A: Tops to Shoes}\label{sec:FMNIST03to59}
Table \ref{tab:exp-FMNIST03to59} shows the results of using images
of dresses and shirts from the FMNIST dataset as the source domain, with FMNIST images of sandals and ankle boots as the target. Results show a best performance when using \gls{BD} and \gls{DAd} regularization along with \gls{DS} feature embeddings and the autoencoder for preserving the underlying data structure. Just using \gls{BD} regularization had the second best results. Looking more closely at the other components, the two other conditions applying the separate feature extractions along with \gls{BD} were $3^{nd}$ and $4^{th}$ in accuracy. In addition, conditions using only \gls{DAd} regularization or \gls{DAd} along with the \gls{DS} embeddings performed worse than Baseline. These results together suggest that the \gls{BD} along with the \gls{DS} together were responsible for the large transfer accuracy over the baseline. Using the \gls{DS} without the appropriate distribution regularization is not sufficient for transfer.

% dress/tshirt  (3, 0) source
% sandal/ ankle boot (5, 9)
\begin{table}[H]
    \small
	\centering
	\begin{tabular}{lcc}
		Experiment Type & Source Accuracy & Target Accuracy \\
		\hline
        % 2020SPIE_FMNIST_30to59
        Baseline (3 dims)		           & 94.1\%$\pm$1.9\%  & 65.3\%$\pm$5.0\% \\
        Domain Adversarial (DAd) (3 dims)  & 95.8\%$\pm$0.7\%  & 57.6\%$\pm$7.5\% \\
        Bregman Divergence(BD) (3 dims)	   & 91.3\%$\pm$2.1\%  & \emph{80.7\%$\pm$9.2\%} \\
        Auto-Encoder (AE) (3 dims)	       & 95.1\%$\pm$1.7\%  & 58.5\%$\pm$5.8\% \\
        DA, AE (3 dims)		               & 95.6\%$\pm$0.6\%  & 58.7\%$\pm$5.1\% \\
        BD, AE (3 dims)	                   & 94.2\%$\pm$1.8\%  & 76.1\%$\pm$5.4\% \\
        Direct Sum (DS) (3 dims)           & 95.6\%$\pm$0.9\%  & 52.8\%$\pm$4.1\% \\
        DS, DA (3 dims)		               & 95.4\%$\pm$1.3\%  & 50.1\%$\pm$0.2\% \\
        DS, BD (3 dims)		               & 96.3\%$\pm$0.5\%  & \emph{79.9\%$\pm$15.4\%} \\
        DS, DA, AE (3 dims)	               & 95.7\%$\pm$0.5\%  & 64.0\%$\pm$4.9\% \\
        DS, BD, AE (3 dims)	               & 96.2\%$\pm$0.3\%  & \emph{79.6\%$\pm$8.7\%} \\
        Everything* (3 dims)               & 95.5\%$\pm$1.0\%  & {\bf 88.0\%$\pm$3.1\%} \\
	\end{tabular}
	\caption{FMNIST Transfer Task: source images are of dresses and shirts;
		target images are of sandals and ankle boots. All experiments used a 3-dimensional latent layer}
	\label{tab:exp-FMNIST03to59}
\end{table}

\subsection{Exp 1B: Shoes to Tops}\label{sec:FMNIST59to03}
Table~\ref{tab:exp-FMNIST59to03} shows the results of
reversing the source and target domains. In this case, the \gls{BD} again lead to a large accuracy improvement over the baseline. However, this was the case without needing the \gls{DS} feature embeddings. Using the \gls{DS} feature embeddings along with \gls{BD} or \gls{DAd} regularization improved transfer accuracy over the baseline, but not as much as just using \gls{BD} regularization. These results together suggest that for this experiment scenario, a common feature embedding could be found to match the source and target distributions. More work would need to be done to identify the particular important features.
% sandal/ ankle boot (5, 9) source
% dress/tshirt  (3, 0)
\begin{table}[H]
	\centering
    \small
	\begin{tabular}{lcc}
		Experiment Type & Source Accuracy & Target Accuracy \\
		\hline
        % 2020SPIE_FMNIST_59to30
		Baseline                 & 98.1\%$\pm$0.8\%  & 56.7\%$\pm$5.1\% \\
		Domain Adversarial (DA)  & 98.7\%$\pm$0.4\%  & 55.3\%$\pm$4.8\% \\
		Bregman Divergence(BD)   & 98.5\%$\pm$0.7\%  & \emph{75.1\%$\pm$2.5\%} \\
		Auto-Encoder (AE)        & 99.1\%$\pm$0.2\%  & 57.2\%$\pm$7.3\% \\
		DA, AE                   & 99.1\%$\pm$0.2\%  & 56.8\%$\pm$9.9\% \\
		BD, AE                   & 98.6\%$\pm$0.4\%  & {\bf 75.7\%$\pm$1.9\%} \\
		Direct Sum (DS)          & 99.1\%$\pm$0.4\%  & 51.2\%$\pm$1.5\% \\
		DS, DA                   & 99.1\%$\pm$0.3\%  & 55.4\%$\pm$5.8\% \\
		DS, BD                   & 99.1\%$\pm$0.2\%  & 66.3\%$\pm$5.7\% \\
		DS, DAd, AE               & 99.1\%$\pm$0.3\%  & 69.6\%$\pm$14.7\% \\
		DS, BD, AE               & 99.3\%$\pm$0.1\%  & 67.4\%$\pm$3.4\% \\
		Everything               & 99.1\%$\pm$0.3\%  & 69.8\%$\pm$2.9\% \\
	\end{tabular}
	\caption{FMNIST Transfer Task: source images are of sandals and ankle boots;
		target images are of dresses and shirts. All algorithms used a 3-dimensional latent space.}
	\label{tab:exp-FMNIST59to03}
\end{table}

% \begin{table}[H]
% 	\centering
% 	\begin{tabular}{lcc}
% 		Experiment Type & Source Accuracy & Target Accuracy \\
% 		\hline
% 		Baseline                 & \%$\pm$  & \%$\pm$ \% \\
% 		Domain Adversarial (DA)  & \%$\pm$  & \%$\pm$ \% \\
% 		Bregman Divergence(BD)   & \%$\pm$  & \%$\pm$ \% \\
% 		Auto-Encoder (AE)        & \%$\pm$  & \%$\pm$ \% \\
% 		DA, AE                   & \%$\pm$  & \%$\pm$ \% \\
% 		BD, AE                   & \%$\pm$  & \%$\pm$ \% \\
% 		Direct Sum (DS)          & \%$\pm$  & \%$\pm$ \% \\
% 		DS, DA                   & \%$\pm$  & \%$\pm$ \% \\
% 		DS, BD                   & \%$\pm$  & \%$\pm$ \% \\
% 		DS, DA, AE               & \%$\pm$  & \%$\pm$ \% \\
% 		DS, BD, AE               & \%$\pm$  & \%$\pm$ \% \\
% 		Everything               & \%$\pm$  & \%$\pm$ \% \\
% 	\end{tabular}
% 	\caption{FMNIST Transfer Task: source images are of ;
% 		target images are of .}
% 	\label{tab:exp-BCTtoPVUV}
% \end{table}

\subsection{Exp 2A: Cross Digit (01-23) and Cross Class (MNIST-USPS)}\label{sec:MNISTtoUSPS0123}
The results of the cross-dataset transfer task consisting
of images of 0's and 1's from MNIST as the source domain and images of
2's and 3's from the USPS dataset can be found in Table \ref{tab:MNISTtoUSPS0123}. In this experiment we also experimented with the dimensionality of the latent space by using both a 3 and 10-dimensional latent space. This allowed us to investigate the effect of varying the manifold dimension. The intuition is that a higher dimensional latent space can represent more class variability, but also requires more parameters to learn. However, higher fidelity representations do not necessary imply better class discrimination. As in the previous ablation studies, the results shown are based on running 5 Monte Carlo iterations for each experiment. Overall, the smaller 3-dimensional latent space showed the best transfer performance. The architecture applying the \gls{DS} embeddings with both \gls{BD} and \gls{DAd} regularization with the autoencoder did the best, while removing just \gls{DAd} regularization from the best architecture only led to a small performance drop (97.9\% to 96.9\% accuracy). As was found in previous experiments, \gls{BD} regularization without the \gls{DS} embeddings also gave performance well above the baseline. These results are consistent with the previous, and show that the \gls{DS} embeddings give a performance boost above a single embedding with divergence minimization, but the particular divergence minimization is important.  \Gls{BD} regularization
again shows an advantage over \gls{DAd} regularization. Furthermore, for this two-class transfer case we found a clear advantage using a smaller latent space.

\begin{table}[H]
	\centering
    \small
	\begin{tabular}{lcccc}
		Experiment Type & Source Acc (3 dim) & Target Acc (3 dim) & Source Acc (10 dim) & Target Acc (10 dim) \\
		\hline
        % 2020SPIE_MNISTtoUSPS_01to23
Baseline 		  & 97.7\%$\pm$1.4\%  & 73.6\%$\pm$12.6\%    & 99.5\%$\pm$0.2\%  & 89.5\%$\pm$6.2\%  \\
Domain Adversarial (DA)   & 97.1\%$\pm$3.4\%  & 82.4\%$\pm$14.5\%  & 98.3\%$\pm$1.4\%  & 82.8\%$\pm$11.4\% \\
Bregman Divergence(BD)    & 94.5\%$\pm$1.8\%  & 94.6\%$\pm$2.6\%  & 98.8\%$\pm$0.7\%  & 91.4\%$\pm$8.0\% \\
Auto-Encoder (AE) 		  & 99.1\%$\pm$0.6\%  & 65.4\%$\pm$9.4\%   & 97.2\%$\pm$2.7\%  & 74.4\%$\pm$16.5\% \\
DA, AE 		  & 97.7\%$\pm$3.2\%  & 69.2\%$\pm$10.6\%   & 98.4\%$\pm$1.0\%  & 80.6\%$\pm$10.4\% \\
BD, AE 		  & 95.4\%$\pm$2.6\%  & \emph{ 95.5\%$\pm$2.4\%}   & 97.0\%$\pm$1.9\%  & 92.2\%$\pm$4.3\% \\
Direct Sum (DS) 		  & 98.2\%$\pm$2.0\%  & 65.8\%$\pm$17.1\%  & 99.1\%$\pm$0.3\%  & 66.8\%$\pm$12.8\% \\
DS, DA 		  & 99.2\%$\pm$0.4\%  & 63.6\%$\pm$17.4\%   & 99.1\%$\pm$0.3\%  & 63.7\%$\pm$17.7\% \\
DS, BD 		  & 98.7\%$\pm$0.3\%  & 92.0\%$\pm$10.7\%   & 97.7\%$\pm$1.9\%  & 75.7\%$\pm$13.6\% \\
DS, DA, AE 		  & 97.8\%$\pm$2.0\%  & 65.8\%$\pm$17.3\%  & 98.5\%$\pm$0.6\%  & 66.0\%$\pm$14.4\% \\
DS, BD, AE 		  & 98.3\%$\pm$1.0\%  & \emph{96.9\%$\pm$2.5\%}   & 98.7\%$\pm$0.8\%  & 78.9\%$\pm$19.9\% \\
Everything 		  & 97.9\%$\pm$1.0\%  & {\bf 97.9\%$\pm$0.8\% } & 98.0\%$\pm$1.0\%  & 73.5\%$\pm$15.0\% \\
	\end{tabular}
	\caption{MNIST to USPS cross dataset transfer task for digits 0 and 1 to 2 and 3. We experimented with both
		a 3 and 10-dimensional latent space.}
	\label{tab:MNISTtoUSPS0123}
\end{table}

\subsection{Exp 2B: Cross Digit (45-39) and Cross Class (USPS-MNIST)}\label{sec:USPStoMNIST45to39}
Reversing the study of the previous cross dataset transfer
task and trying different number classes, we take images of 4's and 5's from the USPS dataset as the source
domain and images of 3's and 9's from MNIST as the target. The results
of this study are shown in Table \ref{tab:USPStoMNIST4539}. In this experiment, unlike the previous digits experiment of Sec.~\ref{sec:MNISTtoUSPS0123}, we found that a simpler architecture with just a single feature embedding and \gls{BD} with or without the autoencoder gave the best transfer performance. Furthermore, unlike the previous experiment, the larger 10-dimensional latent space worked the best.
\begin{table}[H]
	\centering
    \small
	\begin{tabular}{lcccc}
		Experiment Type & Source Acc (3 dim) & Target Acc (3 dim) & Source Acc (10 dim) & Target Acc (10 dim) \\
		\hline
        % 2020SPIE_USPStoMNIST_45to39
Baseline 		  & 99.5\%$\pm$0.6\%  & 82.5\%$\pm$3.6\%  & 99.5\%$\pm$0.3\%  & 79.7\%$\pm$7.8\% \\
Domain Adversarial (DA) 		  & 99.3\%$\pm$0.9\%  & 81.9\%$\pm$4.4\%  & 98.7\%$\pm$1.1\%  & 86.8\%$\pm$2.6\% \\
Bregman Divergence(BD) & 99.5\%$\pm$0.4\%  & \emph{ 92.2\%$\pm$1.2\% }  & 99.6\%$\pm$0.3\%  & {\bf 93.4\%$\pm$0.8\% }\\
Auto-Encoder (AE) 		  & 98.4\%$\pm$1.5\%  & 87.1\%$\pm$2.0\%  & 99.2\%$\pm$0.8\%  & 88.4\%$\pm$1.8\% \\
DA, AE 		  & 99.8\%$\pm$0.1\%  & 86.3\%$\pm$6.3\%  & 99.4\%$\pm$0.5\%  & 88.1\%$\pm$2.2\% \\
BD, AE 		  & 99.3\%$\pm$0.7\%  & 92.1\%$\pm$1.3\%   & 99.4\%$\pm$0.4\%  & {\bf 93.4\%$\pm$1.0\% } \\
Direct Sum (DS)  & 99.4\%$\pm$0.8\%  & 52.7\%$\pm$3.0\%  & 99.5\%$\pm$0.4\%  & 53.8\%$\pm$2.2\% \\
DS, DA 		  & 99.7\%$\pm$0.2\%  & 51.4\%$\pm$0.1\%  & 99.1\%$\pm$1.0\%  & 53.9\%$\pm$5.2\% \\
DS, BD 		  & 99.4\%$\pm$0.4\%  & 81.5\%$\pm$9.4\%  & 99.4\%$\pm$0.5\%  & 67.3\%$\pm$15.3\% \\
DS, DA, AE 		  & 99.8\%$\pm$0.1\%  & 55.2\%$\pm$5.9\% & 99.5\%$\pm$0.2\%  & 53.9\%$\pm$4.9\% \\
DS, BD, AE 		  & 99.5\%$\pm$0.3\%  & 83.9\%$\pm$8.7\%  & 99.0\%$\pm$0.9\%  & 53.7\%$\pm$4.8\% \\
Everything 		  & 99.4\%$\pm$0.3\%  & 77.7\%$\pm$13.1\%  & 98.8\%$\pm$1.0\%  & 61.4\%$\pm$16.4\% \\
	\end{tabular}
	\caption{USPS to MNIST cross dataset transfer task for digits 4 and 5 to 3 and 9. We experimented with both a 3 and 10-dimensional latent space.}
	\label{tab:USPStoMNIST4539}
\end{table}

\subsection{Experiment 3: 10-digit transfer (MNIST and USPS)}\label{sec:MNISTtoUSPS}
Table \ref{tab:USPStoMNIST-MNISTtoUSPS} shows the results of 10 class digit transfer. Surprisingly, on the MNIST to USPS transfer, the \gls{BD} regularization along with the autoencoder only very slightly improved accuracy over the baseline. This is in contrast to the USPS to MNIST transfer, where \gls{BD} regularization led to a 10\% absolute improvement over the baseline.
Another initially surprising result was that dramatic reduction in performance when using the \gls{DS} embedding. We inspected confusion matrices to more precisely understand the error source. The confusion matrices demonstrated that most samples were classified into a small subset of classes. This reveals that the network had learned to \emph{collapse} multiple classes, an issue called ``mode collapse'' in the literature. This results from initializing the network with random weights and not having any labels from the target set to separate classes. It also highlights the utility of initializing the target feature extraction network with the source network as in \gls{ADDA}\cite{tzeng2017adversarial}.

\begin{table}[H]
	\centering
    \small
	\begin{tabular}{lcccc}
		Experiment Type & Source Acc (U$\rightarrow$M)  & Target Acc (U$\rightarrow$M) & Source Acc (M$\rightarrow$U) & Target Acc (M$\rightarrow$U)\\
		\hline
        % 2020SPIE_USPStoMNIST
Baseline  & 89.9\%$\pm$2.4\%  & 47.3\%$\pm$3.6\%  & 87.3\%$\pm$2.2\%  & 73.1\%$\pm$2.0\%\\
Domain Adversarial (DA)   & 88.2\%$\pm$3.1\%  & 50.6\%$\pm$2.0\%  & 81.8\%$\pm$5.0\%  & 64.2\%$\pm$5.0\%\\
Bregman Divergence(BD)  & 89.8\%$\pm$5.9\%  & {\bf 57.8\%$\pm$2.1\% } & 84.5\%$\pm$3.9\%  & 69.0\%$\pm$8.5\%\\
Auto-Encoder (AE)(10 dims)	 & 90.8\%$\pm$0.9\%  & 49.4\%$\pm$2.1\%  & 86.8\%$\pm$3.6\%  & 68.0\%$\pm$4.3\% \\
DA, AE  & 88.4\%$\pm$4.0\%  & 46.2\%$\pm$2.9\%   & 88.5\%$\pm$2.4\%  & 69.8\%$\pm$3.8\% \\
BD, AE   & 89.1\%$\pm$2.4\%  & 55.2\%$\pm$3.7\%  & 80.7\%$\pm$9.2\%  & {\bf 73.2\%$\pm$2.5\% }\\
Direct Sum (DS)(10 dims)	 & 91.5\%$\pm$3.2\%  & 8.5\%$\pm$2.2\%  & 88.3\%$\pm$2.7\%  & 8.2\%$\pm$1.2\%\\
DS, DA  & 91.6\%$\pm$3.3\%  & 9.4\%$\pm$0.6\%  & 87.6\%$\pm$2.4\%  & 7.8\%$\pm$1.8\% \\
DS, BD    & 91.5\%$\pm$1.7\%  & 8.3\%$\pm$2.4\%   & 87.7\%$\pm$1.9\%  & 9.3\%$\pm$5.0\%\\
DS, DA, AE   & 91.9\%$\pm$1.3\%  & 10.4\%$\pm$1.5\%   & 84.9\%$\pm$9.8\%  & 9.5\%$\pm$4.0\% \\
DS, BD, AE   & 90.8\%$\pm$3.0\%  & 13.0\%$\pm$4.1\%  & 87.6\%$\pm$1.2\%  & 13.7\%$\pm$5.4\%  \\
Everything  & 93.1\%$\pm$1.1\%  & 14.9\%$\pm$1.9\%   & 84.7\%$\pm$6.0\%  & 8.2\%$\pm$3.8\%\\
	\end{tabular}
	\caption{USPS (U) to MNIST (M) on left and M to U on right. All experiments used a 10-dimensional latent space.}
    \label{tab:USPStoMNIST-MNISTtoUSPS}
\end{table}

% -----------------------------
\subsection{Experiment 4: Synthetic to Measured SAR}\label{sec:SAMPLE}
For the next transfer task, we consider the synthetic-to-measured transfer
task where our source domain consists of 10 classes of synthetic SAR images from
the SAMPLE dataset\cite{SAMPLE} and our target domain consists of corresponding measured SAR images. The results are shown in Table \ref{tab:SAMPLE}. Much like in the 10-category experiment of Sec.~\ref{sec:MNISTtoUSPS}, \gls{DS} embeddings caused mode collapse and poor performance for transfer learning. However, unlike the previous experiments, \gls{DAd} outperformed \gls{BD} regularization, with \gls{BD} regularization only narrowly outperforming the baseline. Best performance was achieved by a single autoencoder with \gls{DAd} regularization. The simple autoencoder did nearly as well (40.1\% versus 37.7\% accuracy).

\begin{table}[H]
	\centering
	\begin{tabular}{lcc}
		Experiment Type & Source Accuracy & Target Accuracy \\
		\hline
        % 2020SPIE_SAMPLE_SynthToMeas
		Baseline 	  & 85.7\%$\pm$9.9\%  & 31.4\%$\pm$5.4\% \\
		Domain Adversarial (DA) 	  & 84.1\%$\pm$6.0\%  & 31.7\%$\pm$8.1\% \\
		Bregman Divergence(BD) 	  & 67.8\%$\pm$19.7\%  & 33.2\%$\pm$5.1\% \\
		Auto-Encoder (AE)(10 dims)		  & 61.7\%$\pm$16.5\%  & 37.7\%$\pm$9.3\% \\
		DA, AE 	  & 78.4\%$\pm$8.8\%  & {\bf 40.1\%$\pm$3.2\%} \\
		BD, AE 	  & 49.5\%$\pm$14.7\%  & 33.3\%$\pm$12.0\% \\
		Direct Sum (DS)(10 dims)		  & 78.8\%$\pm$11.1\%  & 10.3\%$\pm$2.2\% \\
		DS, DA 	  & 82.6\%$\pm$17.4\%  & 10.2\%$\pm$1.3\% \\
		DS, BD 	  & 92.3\%$\pm$8.0\%  & 11.3\%$\pm$3.8\% \\
		DS, DA, AE 	  & 80.7\%$\pm$15.1\%  & 8.5\%$\pm$4.4\% \\
		DS, BD, AE 	  & 89.8\%$\pm$5.8\%  & 10.7\%$\pm$3.1\% \\
		Everything 	  & 79.7\%$\pm$12.1\%  & 13.3\%$\pm$0.9\% \\
	\end{tabular}
	\caption{SAMPLE Synthetic to Measured Cross Dataset Transfer Task. All experiments used a 10-dimensional latent space.}
	\label{tab:SAMPLE}
\end{table}

% -----------------------------
\subsection{Experiment 5: Cross Vehicle Satellite Imagery}\label{sec:xView}
In the next experiment of Table \ref{tab:xView} we consider a two cross category transfer experiment on the xView satellite imagery dataset\cite{xView}. As in the Experiments of Rivera \etal\cite{Rivera2019}, for source we used 'bus' and 'cargo truck' while we used 'passenger vehicle' and 'utility truck' as the target classes. As with previous 2 cross class transfer experiments, the \gls{DS} embedding gave a slight performance edge over the single feature embedding. However, the results are only slightly better than chance while being better than the baseline.

\begin{table}[H]
	\centering
	\begin{tabular}{lcc}
		Experiment Type & Source Accuracy & Target Accuracy \\
		\hline
        % 2020SPIE_XVIEW_bcToPu
		Baseline (3 dims)		  & 75.6\%$\pm$3.1\%  & 52.1\%$\pm$0.6\% \\
		Domain Adversarial (DA) (3 dims)		  & 78.0\%$\pm$1.0\%  & 52.9\%$\pm$1.1\% \\
		Bregman Divergence(BD) (3 dims)		  & 66.6\%$\pm$9.3\%  & 53.5\%$\pm$1.9\% \\
		Auto-Encoder (AE)(3 dims)		  & 74.1\%$\pm$2.2\%  & 52.6\%$\pm$1.5\% \\
		DA, AE  & 74.6\%$\pm$5.8\%  & 53.0\%$\pm$1.1\% \\
		BD, AE (3 dims)		  & 64.2\%$\pm$5.3\%  & 53.7\%$\pm$3.2\% \\
		Direct Sum (DS)(3 dims)		  & 73.8\%$\pm$7.6\%  & 51.6\%$\pm$3.0\% \\
		DS, DA (3 dims)		  & 76.5\%$\pm$4.1\%  & 52.6\%$\pm$3.1\% \\
		DS, BD (3 dims)		  & 74.0\%$\pm$2.9\%  & 54.9\%$\pm$3.8\% \\
		DS, DA, AE (3 dims)		  & 72.7\%$\pm$4.1\%  & 54.4\%$\pm$3.9\% \\
		DS, BD, AE (3 dims)		  & 74.3\%$\pm$2.1\%  & 55.5\%$\pm$1.9\% \\
		Everything (3 dims)		  & 71.3\%$\pm$6.9\%  & 54.3\%$\pm$2.8\% \\
	\end{tabular}
	\caption{xView Transfer Task. All experiments used a 3-dimensional latent space.}
	\label{tab:xView}
\end{table}

\section{DISCUSSION}
\label{sec:discussion}  % \label{} allows reference to this section

The results of Sec.~\ref{sec:experiments} show a few consistent patterns:
\begin{enumerate}
\item Separate feature extractions supported better transfer across two classes in some cases, but multi-class sets favored a single feature extractor,

\item In all cases except for the synthetic-to-measured experiment, \gls{BD} outperformed \gls{DAd} regularization.
\end{enumerate}
We discuss those patterns below.

\subsection{Separate versus Single Feature Extraction}
One of the key contributions of this work is to show that by treating the source and target data as existing in separate feature spaces, we can transfer more flexibly across object classes than in the baseline case. We demonstrated this in the experiment of Sec.~\ref{sec:MNISTtoUSPS0123}, where we achieved 97.9\% accuracy on both source and target recognition for a two-digit transfer task where the baseline achieved 73.6\%. However, this was not always the case. In some cases, a single feature common extraction gave better performance, but it typically was coupled with either \gls{BD} regularization or the autoencoder for preserving data structure. The \gls{DS} shortcoming becomes increasingly evident for the multiple class transfer experiments where mode collapse caused chance performance for all architectures using the \gls{DS} feature extraction. Taken together, the results show that learning separate feature embeddings can provide increased flexibility, but comes at the cost of being more challenging to train. It may require a different type of training procedure or some supervision. Such questions pose directions for future work.

\subsection{\Gls{BD} versus \gls{DAd} Regularization}
Another important contribution is the application of \gls{BD} within this network learning framework. Unlike the popular \gls{DAd} regularization that acts as a proxy for the difference between source and target distributions, the \gls{BD} explicitly penalizes the distribution differences. The experiments overwhelmingly favored \gls{BD}, except for the synthetic-to-measured study in Sec.~\ref{sec:SAMPLE}. The drawback of \gls{BD} is the increased computational burden of \gls{KDE} and the associated gradient calculation. Fortunately, the burden happens during training so it can be done offline. \Gls{DAd} regularization also has training drawbacks and is prone to vanishing or exploding gradients. In our preliminary experiments we found that batch normalization helped with the vanishing gradient problem. We also found that values larger than $\alpha_{\text{DA}}=0.1$ destabilized network training.

\section{CONCLUSION}
\label{sec:conclusion}  % \label{} allows reference to this section
We have presented the \gls{DiSDAT} network for transfer learning from a labeled source dataset to an unlabeled target dataset. The key idea is to learn separate feature embeddings to a common latent space by conceptualizing the source and target as existing in separate  regions of a combined source and target direct sum space. By taking this view, we can more flexible transfer across different object classes and feature spaces. The important constraint is that we must match the target data to the source data in the latent space through appropriate regularization. Our experiments show that minimizing the distribution divergence while preserving the lower dimensional data manifold gives good results. The major drawback is mode collapse which occurs for multi-class sets and will be addressed with future work.

%%%%%%%%%%%%%%%%%%%%%%%%%%%%%%%%%%%%%%%%%%%%%%%%%%%%%%%%%%%%%

%%%%%%%%%%%%%%%%%%%%%%%%%%%%%%%%%%%%%%%%%%%%%%%%%%%%%%%%%%%%%
\acknowledgments     %>>>> equivalent to \section*{ACKNOWLEDGMENTS}

We would like to thank Dr. Olga Mendoza-Schrock and Mr. Christopher Menart for their input and feedback during this research project. This work was supported by Air Force Research Laboratory (AFRL), Air Force Office of Scientific Research (AFOSR), Dynamic Data Driven Application Systems (DDDAS) Program, Autonomy Technology Research Center (ATRC), and Dr. Erik Blasch.

% %%%%%%%%%%%%%%%%%%%%%%%%%%%%%%%%%%%%%%%%%%%%%%%%%%%%%%%%%%%%%
% %%%%% TODO Notes %%%%%
%
% \newpage
% \listoftodos[TODO Notes]

%%%%%%%%%%%%%%%%%%%%%%%%%%%%%%%%%%%%%%%%%%%%%%%%%%%%%%%%%%%%%
%%%%% References %%%%%

\bibliography{rivera}   %>>>> bibliography data in report.bib
\bibliographystyle{spiebib}   %>>>> makes bibtex use spiebib.bst

\end{document}